\documentclass[runningheads]{llncs}

\usepackage{multirow}
\usepackage{eccv}

\usepackage{eccvabbrv}

\usepackage{graphicx}
\usepackage{booktabs}

\usepackage[accsupp]{axessibility}  

\usepackage[pagebackref,breaklinks,colorlinks,citecolor=eccvblue]{hyperref}

\usepackage{orcidlink}

\begin{document}

\title{Enhancing Weakly-Supervised Histopathology Image Segmentation with Knowledge Distillation on MIL-Based Pseudo-Labels}

\titlerunning{Abbreviated paper title}

\author{Yinsheng He \and
Xingyu Li \and
Roger J. Zemp}


\institute{University of Alberta}

\maketitle

\begin{abstract}

Segmenting tumors in histological images is vital for cancer diagnosis. While fully supervised models excel with pixel-level annotations, creating such annotations is labor-intensive and costly. Accurate histopathology image segmentation under weakly-supervised conditions with coarse-grained image labels is still a challenging problem. Although multiple instance learning (MIL) has shown promise in segmentation tasks, surprisingly, no previous pseudo-supervision methods have used MIL-based outputs as pseudo-masks for training. We suspect this stems from concerns over noises in MIL results affecting pseudo supervision quality. To explore the potential of leveraging MIL-based segmentation for pseudo supervision, we propose a novel distillation framework for histopathology image segmentation. This framework introduces a iterative fusion-knowledge distillation strategy, enabling the student model to learn directly from the teacher's comprehensive outcomes. Through dynamic role reversal between the fixed teacher and learnable student models and the incorporation of weighted cross-entropy loss for model optimization, our approach prevents performance deterioration and noise amplification during knowledge distillation. Experimental results on public histopathology datasets, Camelyon16 and Digestpath2019, demonstrate that our approach not only complements various MIL-based segmentation methods but also significantly enhances their performance. Additionally, our method achieves new SOTA in the field. Codes are available at https://github.com/wollf2008/IKD-MIL.
\end{abstract}

\section{Introduction}
\label{sec:intro}

In the current medical system, the visual examination of histological images is known as the gold standard for the identification of cancer. Accurate identifying cancerous regions 
are a critical prerequisite for pathologists in determining cancer invasiveness. Over the past decade, a multitude of deep learning-based algorithms have been developed for this purpose. These methods mainly fall into three categories: supervised~\cite{munoz2020pyhist}~\cite{he2023whole}, weakly supervised~\cite{campanella2019clinical}~\cite{xu2014weakly}, and unsupervised~\cite{kumar2017dataset}~\cite{bulten2018unsupervised}. In recent years, with significant advancements in deep learning approaches, the performance of fully supervised learning networks has approached, and in some aspects, even close to that of human experts~\cite{nagpal2019development}. However, such success is largely attributable to the availability of extensive, high-quality manually annotated data. To obtain sufficiently accurate ground truth, pathologists typically spend considerable time annotating histopathological images with ultra-fine granularity, consuming substantial labor costs. On the other hand, unsupervised learning strategies, which require no annotations whatsoever, are still a long way from practical application due to lower accuracy. Weakly supervised learning, as a strategy that necessitates only coarse annotations, achieves a better balance in terms of cost and performance~\cite{qian2022transformer}.

\begin{figure}[t]
    \centering
    \includegraphics[scale = 0.43]{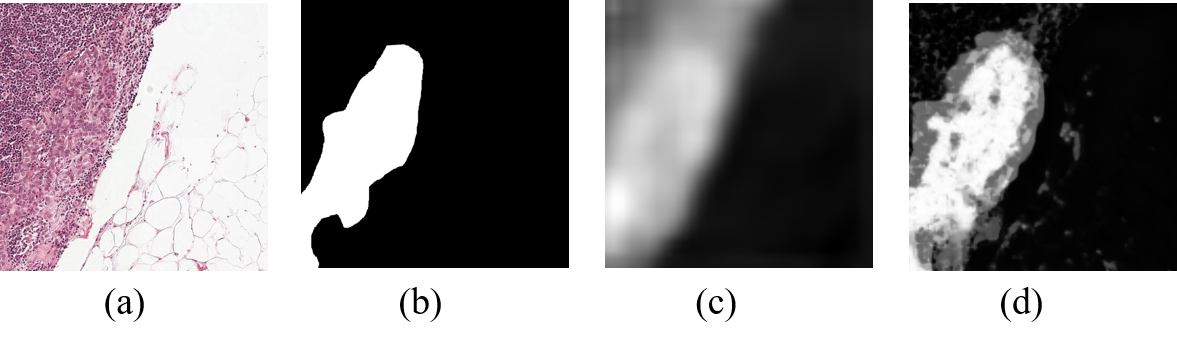}
    \caption{Pseudo-mask comparison between MIL method and CAM method. (a) original image, (b) ground truth mask for the image, (c) pseudo-mask generated by CAM method, (d) pseudo-mask generated by MIL method. The challenge of using MIL-based segmentation as pseudo masks lies in how to prevent the segmentation model from deteriorating over time under pseudo supervision.}
    \label{fig:MILvsCAM}
\end{figure}

Weakly supervised segmentation allows for the utilization of coarse-grained, image-level annotations as guidance. In this process, pathologists merely indicate the presence or absence of cancerous tissues within training images; the model then proceeds to autonomously conduct pixel-level segmentation on new data. Multiple instance learning (MIL)~\cite{dietterich1997solving} has been exploited for this purpose. It consider pixels as instances within a bag (i.e. an image) and infers the specific pixel-level labels from the bag label~\cite{simonyan2014very, he2016deep, li2023weakly,qian2022transformer,lin2023interventional}. Alternatively, pseudo-supervision serves as another prevalent method for addressing this task. It leverages coarse-grained annotation information to generate pseudo-masks which are then used as targets for training a segmentation model in a manner akin to supervised learning. In prior arts, class Activation Mapping (CAM) is commonly used to produce these pseudo-masks~\cite{han2022multi, jiang2019integral, li2022online,chen2022c,fang2023weakly,pan2023cvfc,kuang2024weakly}. Nevertheless, CAM's inability to accurately define object boundaries often results in pseudo-masks that are both blurred and imprecise~\cite{ahn2018learning}. Consequently, CAM-based pseudo-supervision may not consistently apply identical labels to the same anatomical structures across various histopathology images, highlighting a potential limitation in achieving precise segmentation~\cite{chaitanya2023local}. 
In theory, both CAM-generated and MIL-derived segmentation results hold potential for use as pseudo-masks within the pseudo-supervision framework for histopathology image segmentation. As demonstrated in Fig.~\ref{fig:MILvsCAM}, segmentation outcomes from MIL have higher granularity and more defined boundaries compared to the typically blurred pseudo-masks produced by CAM-based methods, albeit potentially introducing more noise~\cite{li2023weakly,qian2022transformer}. Intriguingly, a review of existing literature reveals an absence of studies employing MIL-based outcomes as pseudo-masks for training. This observation leads us to hypothesize that the direct integration of these noise-prone MIL-based segmentation results as pseudo-masks might adversely affect the segmentation model's accuracy over time due to pseudo supervision.

\begin{figure}[t]
    \centering
    \includegraphics[scale = 0.34]{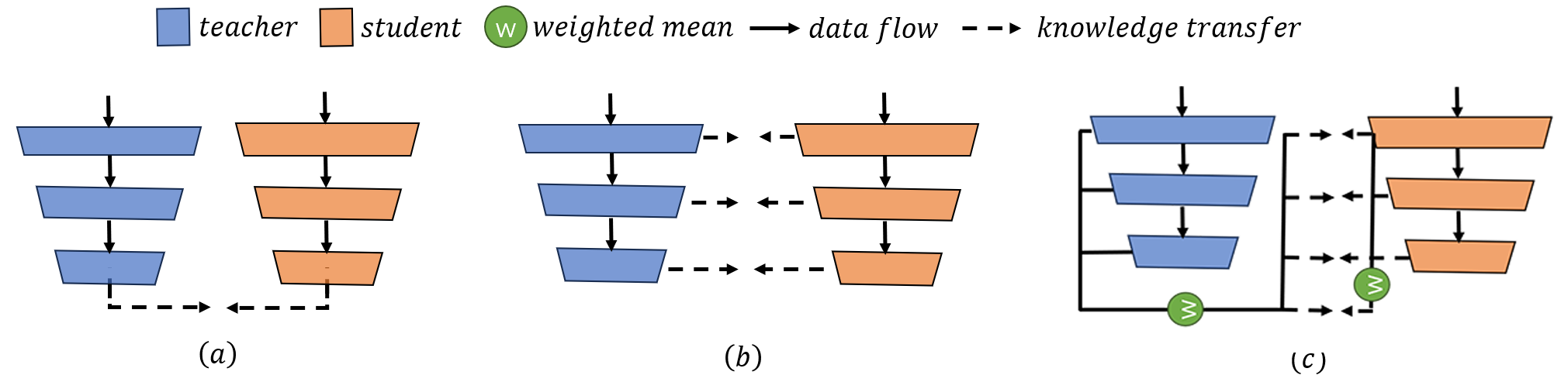}
    \caption{Knowledge distillation structures, (a), (b) are common knowledge distillation methods, and (c) is our fusion knowledge distillation. The comparison between these methods is shown in Fig.~\ref{fig:Camelyon16_fusion} and Fig.~\ref{fig:Digestpath2019_fusion}.}
    \label{fig:skd}
\end{figure}

To verify our hypothesis and fill the existing research void, this study investigates MIL-based pseudo-supervision for weakly supervised segmentation in histopathology images. To mitigate the potential degradation in model performance due to pseudo supervision, we introduce a novel knowledge distillation approach designed to both stabilize and enhance the segmentation model. Specifically, we innovate a iterative fusion-knowledge distillation strategy to address the noise and label-ambiguity issue in MIL. We show that this strategy can be seamlessly integrated with a variety of MIL-based segmentation architectures, leading to notable improvements in performance. As demonstrated in Fig.~\ref{fig:skd}, diverging from the common knowledge distillation methods which align the student model's behavior with that of the teacher at either the final output or intermediate representations, our fusion knowledge distillation guides each block of the student model to directly assimilate the comprehensive insights from teacher's final outcome. Echoing the findings of ~\cite{zhang2021self}, our approach empowers even the shallow layers of the student model to capture features typically reserved for deeper layers, enabling effective segmentation of both local and global cancerous regions. Another unique aspect of our knowledge distillation is the dynamic role reversal between teacher and student models once the student outperforms the teacher, followed by a repeat of the distillation cycle to refine segmentation accuracy further. During each distillation cycle, the fixed (frozen) teacher model enables the student to consistently learn and enhance its capabilities from the teacher's pre-established, static insights. To further counteract the potential magnification of noise during knowledge distillation, we incorporate a weighted cross-entropy loss~\cite{li2022online} as a regularization mechanism in the optimization of the student model. Our experimental evaluation across two publicly available histopathology datasets confirms that our proposed methodology sets a new benchmark for state-of-the-art segmentation performance. 

\begin{itemize}
\item We introduce a MIL-based pseudo-supervision framework for weakly supervised segmentation with coarse-grained labels on histopathology images. This framework can be seamlessly integrated with a variety of MIL-based segmentation models for performance boost, opening a new avenue of leveraging MIL-based results in the pseudo-supervision paradigm. 

\item We innovate a fusion-knowledge distillation strategy. It empowers shallow layers of the student model to capture abstract features typically associated with deeper layers, enhancing segmentation performance for both local and global regions of interest. Special designs such as dynamic teacher-student switch help prevent model deterioration over time.   
    
    
    
    \item We conduct extensive experiments to validate the effectiveness of the proposed method and report a new SOTA in the field.
\end{itemize}

\section{Related Work}

\subsection{Histopathology Image Weakly Supervised Segmentation}

\textbf{Pseudo-supervision} is a common strategy in Weakly Supervised Semantic Segmentation (WSSS). 
In the field of histopathology imaging, pseudo-supervision is often tied with Class Activation Mapping (CAM)~\cite{simonyan2014very, he2016deep, zhou2016learning, li2023weakly,qian2022transformer,lin2023interventional,han2022multi, jiang2019integral,chen2022c,fang2023weakly,pan2023cvfc,kuang2024weakly}. The basic idea is to train a classification network and use corresponding CAM images as pseudo-masks, assisting another segmentation network in fully supervised learning~\cite{chan2019histosegnet}. 
It should be noted that the segmentation task aims to identify the complete objects; but CAM usually highlights the most discriminative part, thereby compromising segmentation results. Moreover, due to the high homogeneity of histopathological images, CAM might confuse local activations caused by different representations within an object~\cite{li2022online}. To address these challenges, SCCAM~\cite{chang2020weakly} proposed to sub-categorize objects by feature-level clustering before the pseudo supervision procedure. 
Han \etal~\cite{han2022multi} introduced a Progressive Drop Attention (PDA) method to deactivate highlighted regions gradually. Alternatively, PistoSeg~\cite{fang2023weakly} trained a Synthesized Dataset Generation Module, partially replacing the functionality of CAM.  C-CAM~\cite{chen2022c} introduces a causality module to regularize CAM maps by addressing organ co-occurrence in medical images.

\textbf{Multiple Instance Learning} is also widely adopted for various image segmentation~\cite{li2013harvesting}~\cite{papandreou2015weakly}~\cite{wang2013max}~\cite{pinheiro2015image} and classification tasks~\cite{bontempo2023mil}~\cite{wang2020ud}. It was initially proposed by Dietterich\etal~\cite{dietterich1997solving}, where training is done on sets of instances (referred to as "bags") instead of individual instances. In binary classification settings, a bag is labeled positive if it contains at least one positive instance, otherwise, it's labeled negative. With the rise of neural networks, particularly Convolutional Neural Networks (CNNs)~\cite{lecun1995convolutional} for rich hierarchical feature learning, deep learning have been extensively combined with MIL methods, achieving outstanding results~\cite{xu2014deep}. In the field of histopathological imaging, MIL was first applied to patch-level classification tasks within Whole Slide Images (WSIs)~\cite{xu2012context}. In such tasks, each WSI is treated as a bag, and each patch is regarded as an instance. This approach allowed histopathologists to annotate only at the image level, significantly reducing the labeling time and cost. Subsequently, Jia \etal further proposed treating each labeled histopathological image as a bag and referring image pixels as instances for pixel-level segmentation~\cite{jia2017constrained}. Since then, many emerging techniques have been applied to MIL. 
For instance, Hashimoto \etal~\cite{hashimoto2020multi} incorporated multi-scale feature learning into MIL for WSI segmentation and Ilse \etal~\cite{ilse2018attention} introduced attention mechanisms into MIL to address its blurred boundary issue. 

This research introduces a fusion-knowledge distillation framework designed to counteract potential degradation from noise in MIL outcomes, opening a new avenue of leveraging MIL in the pseudo-supervision paradigm.

\subsection{Knowledge Distillation}
Knowledge distillation~\cite{hinton2015distilling} is a technique that relies on teacher-student (T-S) architecture and targets to transfer knowledge from one model (teacher model) to another (student model). 
As shown in Fig.~\ref{fig:skd}, beyond labels, the student model can learn various feature representations from the teacher model, such as prediction probabilities, feature vectors, and more ~\cite{zhang2021self}. Since the student model usually exhibits superior performance over the teacher model~\cite{furlanello2018born}, methods like momentum update have been widely used to further improve teacher's performance by updating it through the student model~\cite{caron2021emerging}. 
Particularly, in iterative knowledge distillation, the student itself acts as the teacher, leveraging its past predictions for more informative supervision during training for generalization capability enhancement~\cite{kim2020self}. To prevent the occurrence of model collapses in such self-supervised learning tasks, various strategies including contrastive loss~\cite{wu2018unsupervised}, clustering constraints~\cite{caron2020unsupervised}, predictor~\cite{grill2020bootstrap} or batch normalizations~\cite{richemond2020byol} are adopted in model optimization process.  

It's important to highlight that, in contrast to the ensemble teacher distillation approach for classification mentioned in~\cite{kim2020self}, our knowledge distillation method maintains the teacher model unchanged during the student's training phase. Once the student model outperforms the teacher, their roles are reversed, and a new cycle of knowledge distillation begins.

\section{Methodology}

\textbf{Problem formulation.} Let \(X=\{x_1,x_2, ... ,x_n|x_i \in \mathbb{R}^{3 \times H \times W}\}\) be the training data with their image-level labels denoted as $Y = \{y_1,y_2, ... ,y_n|y_i\in\{0,1\}\}$, where \(y_i = 0\) indicates the corresponding image containing normal tissue only, whereas \(y_i = 1\) suggests that \(x_i\) contains either a combination of normal and cancerous tissues or solely cancerous tissue. With the coarse-grained supervision $\{X,Y\}$, the goal of this study is to train a segmentation model that can predict pixel-level masks for histopathology images. 

\begin{figure*}[t]
\centering
\includegraphics[width=0.95\textwidth]{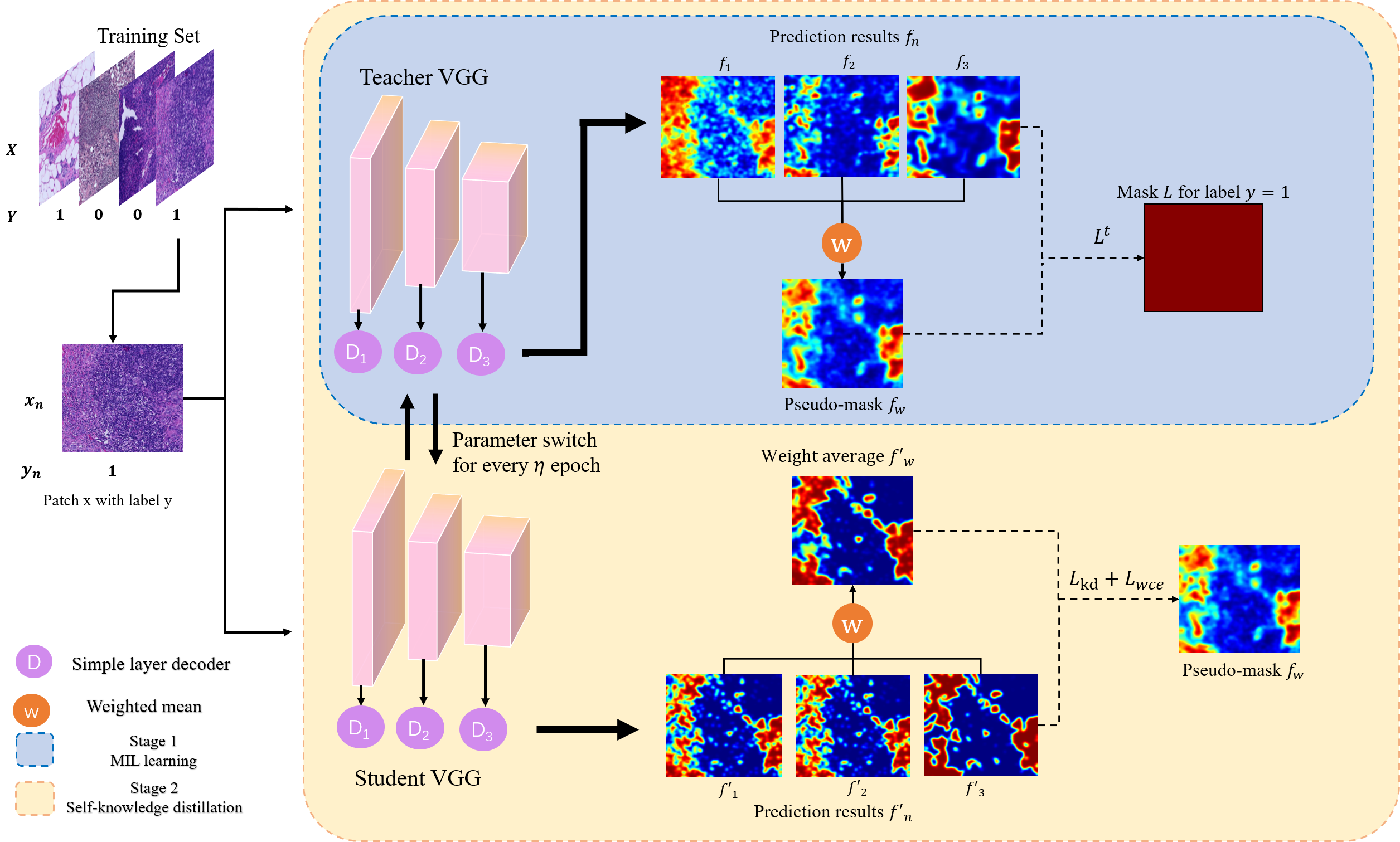}
\caption{An overview of our proposed weakly-supervised segmentation method. Our method consists of two basic stages. In the first stage, we train a simple teacher model through MIL with input image patches and patch labels. In the second stage, we use iterative knowledge distillation to achieve precise pixel-level segmentation. Note, in the knowledge distillation process, the teacher model is frozen before parameter switch.}
\label{fig:1}
\end{figure*}

\subsubsection{System Overview.} To explore MIL for pseudo-supervision in weakly supervised segmentation, a critical hurdle to overcome is to prevent potential adversarial effects of noise present in MIL-derived segmentation outputs. To tackle this challenge, we introduce a iterative fusion-knowledge distillation framework to refine and stabilize the pseudo supervision learning. As illustrated in Fig.~\ref{fig:1}, our MIL-based knowledge distillation framework consists of two training stages. Briefly, we first train a simplistic segmentation model following the MIL paradigm for pixel-level pseudo segmentation masks. To refine the pseudo-mask, we then construct a teacher-student framework and optimize the student segmentation net by a novel self-supervised fusion-knowledge distillation strategy. Particularly, this strategy encourages each block of the student model to align with the teacher's final output directly, facilitating effective segmentation of local and global cancerous regions. 
It is worth noting that though both of the teacher and student models in Fig.~\ref{fig:1} adopt the first three convolutional blocks of VGG16~\cite{simonyan2014very} as the backbones, we show in our experiments that our method can easily adapt to different MIL-based backbones and boost their performance significantly. We will specify the two training stages as follows.

\subsection{MIL for Pseudo-Mask Generation}


In our multiple instance learning, we take each image pixel as one instance and an image as a bag. Following the convention, the initial mask \(l_i\in \mathbb{R}^{\times H \times W}\) of data \(x_i\) for teacher model training are naively set as follows: if \(y_i = 0\), \(l_i = zeros(H, W)\); Otherwise, \(l_i = ones(H, W)\). This naively-initialized training set \(\{X,L\}=\{(x_i,l_i)\}\) is then exploited to train the segmentation model \(T\). Considering the cancerous regions can be any size, we follow the feature pyramid paradigm and let each block generate an independent segmentation probability map through a lightweight 1x1 convolution layer. Then, the output is upscaled to the original image size through bilinear interpolation. 
By minimizing the difference between each of these probability maps and the actual labels, the system effectively discerns and learns the unique characteristics differentiating normal tissues from cancerous ones. 

Mathematically, given an image \(x\), let $\{f_1^t, f_2^t, f_3^t\}$ be its multi-scale segmentation results by the model $T$. We define their aggregated outcome $f_w^t = \sum_i w_if_i^t$ with weight parameter $w$. To determine the optimal value of $w$ and prevent overfitting, we set $w$ as a learnable parameter and conducted a separate training for 10 epochs to ascertain its final value. Then, a compound loss is designed to optimize the teacher segmentation net:
\begin{equation}
L^t =  Dice({f_w^t}, \check{l}) + \sum_{i = 0}^3 Dice({f_i^t}, \check{l}),
\label{eq:teacher dice}
\end{equation}
where \(Dice\) refers to the dice loss. Notably, instead of directly using the naive segmentation mask \(l\) for model optimization, drawing on the optimization approach in~\cite{he2024whole}, we let $\check{f_i^t} = 1-f_i^t$, $\check{f_w^t} = 1-f_w^t$ and $\check{l} = 1-l$, when \(y = 0\); otherwise, $\check{f_i^t} = f_i^t$, $\check{f_w^t} = f_w^t$ and $\check{l} = l$. The reason is that normal samples contain only normal tissue \(y = 0\); thus, the direct application of Dice loss would result in the Dice loss for all positive samples consistently being 1. Consequently, the model would tend to favor the class with a more substantial presence in the dataset, predicting all areas as cancerous regions.


After the MIL training, the segmentation model \(T\) is used as the teacher in the second stage, i.e. knowledge distillation. 

\subsection{Pseudo Supervision with Iterative Fusion-Knowledge Distillation}
Due to the uncertainty and label ambiguity in multiple instance learning, the pseudo-mask generated by the teacher model \(T\) may be noisy and inaccurate, which is likely to receive excessive focus during certain epochs of the training process~\cite{caron2021emerging}. Thus, the challenge of using the results of MIL-based segmentation as pseudo masks lies in how to prevent the segmentation model from deteriorating over time under pseudo supervision. To address this issue, we propose a novel knowledge distillation procedure which is composed by three basic elements: \textit{fusion-knowledge distillation strategy}, (2) \textit{distillation regularization}, and (3) \textit{dynamic teacher-student switch}.


\subsubsection{Fusion-knowledge distillation} distinguishes itself from conventional knowledge distillation approaches by targeting a distinct distillation objective. Its focus shifts from merely mimicking individual blocks in the teacher model to a more direct and holistic learning process. As illustrated in Fig.~\ref{fig:skd}, our fusion knowledge distillation method mandates that each block in the student model directly learns from the final output of the teacher model. This unique strategy involves guiding each block of the student model to acquire a balanced representation of both high-level and low-level features. This innovation empowers even the shallow layers of the student model to capture abstract features typically associated with deeper neural networks. Thereby providing the student network with the opportunity to achieve performance surpassing that of the teacher network.


In specific, for an input image \(x\), the teacher model and student model will each generate their multilevel results, denoted by $\{f_1^t, f_2^t, f_3^t\}$ and $\{f_1^s, f_2^s, f_3^s\}$ respectively. The student follows the same manner to calculate the weighted average $f_w^s$. Then, the fusion-knowledge distillation loss can be formulated as
\begin{equation}
L_{kd} = Dice(\check{f_w^s}, \check{f_w^t}) + \sum_{i = 0}^3 Dice(\check{f_i^s}, \check{f_w^t}).
\label{eq:student dice}
\end{equation}
Note, that since the normal patches only contain normal tissue, we use this ground-truth mask to optimize the student model. Following the process in teacher optimization, when \(y=0\), $ \check{f_i^t} = 1 - f_i^t$, $ \check{f_w^s} = 1 - f_w^s$  and $ \check{f_w^t}=\check{f_i^t} = 0$.

\subsubsection{Distillation regularization} is introduced to prevent performance deterioration in the knowledge distillation process. As shown in Fig.~\ref{fig:MILvsCAM}, MIL-based pseudo-masks invariably contain a substantial amount of noise. These noises may be intensified throughout the self-distillation, thereby imposing a detriment far greater than other methodologies. Specifically, we incorporate a weighted cross-entropy term $L_{wse}$ ~\cite{li2022online} into our distillation process.
\begin{equation}
L_{wce} = \sum_i ce_i * \frac{e^{-ce_i}}{\sum_i e^{-ce_i}},
\label{eq:cross entropy loss}
\end{equation}
where $ce_i=-f_{w,i}^t\log f_{w,i}^s$ is the pixel-level cross-entropy between segmentation predictions from the teacher and student models. This regularization term brings two benefits: (1) it prevents the student model from intensifying noise in pseudo masks during training, and (2) it encourages Teacher-student block-wise feature alignment statistically.

In summary, our knowledge distillation objective function for student training is
\begin{equation}
L^s = L_{kd} + a* L_{wce},
\label{eq:student total loss}
\end{equation}
where $a$ is a scale factor for  $L_{wce}$. If $a$ is small, it fails to significantly impact the model's learning process; conversely, when $a$ is large, it causes the model to overlook smaller cancerous areas. 
In this study, we used for $a$ is 0.25. Ablation on this hyperparameter is shown in Table \ref{table: a ablation}.

\subsubsection{Dynamic Teacher-student Switch} is explored in our iterative knowledge distillation framework. Instead of the slow-paced momentum update on the teacher model during knowledge distillation, we initially freeze the model \(T\) as a static teacher for stable supervision. This effectively counters noise amplification during student model updates by providing a consistent reference, preventing error propagation in knowledge distillation and performance deterioration over time. With the training progress, the student gradually catches up, eventually outperforming the teacher model. Then, we completely swap the role of the two models, turning the original student model into a new teacher, and the original teacher model into the student for the next round of knowledge distillation. This iterative knowledge distillation cycle repeats several times to remove noise and imperfection in the evolving pseudo-masks. We show in the ablations that switching the roles of teacher and student models significantly boost the segmentation performance.

\section{Experiments}
\subsection{Datasets}
We evaluate our approach on two public histopathology datasets as follows. 

\noindent\textbf{Camelyon16}~\cite{bejnordi2017diagnostic} contains 398 annotated Hematoxylin and Eosin (H\&E) stained slides for detecting breast cancer metastasis in sentinel lymph nodes, among which a total of 270 slides were allocated for the training set and the rest 128 slides were for testing. The training set comprised 111 tumor slides and 159 normal slides. The testing set included 80 normal slides and 48 tumor slides. 
These slides are stored in a multi-resolution pyramid. At level 0, each slide has a size of around 100,000 × 100,000 pixels, with x40 magnification (226 nm/pixel). 

To prepare the datasets for our weakly supervised task,  we crop image patches from slides into 1280 * 1280 pixels at level 1 (x20 magnification) in our experiment. Specifically, from the 111 tumor slides in the training set, we dropped those patches with white backgrounds exceeding 80\%, ultimately retaining 6,570 positive (tumor) patches in our training set. We then cropped the same number of image patches from normal slides and included them in our training set, herein referred to as neg-patches.
Similarly, we cropped 985 pos-patches and an equivalent number of neg-patches from the test set. Notably, when selecting the pos-patches for the testing set, we also dropped all patches containing more than 90\% background areas. The rationale behind this is that including an excess of such patches in the test set could unintentionally skew the weakly supervised models' test results towards those of fully supervised models, obscuring the true measure of performance.


\textbf{DigestPath 2019}~\cite{da2022digestpath} is a Colonoscopy tissue segmentation dataset, which contains 250 images of tissue from 93 WSIs, with pixel-level annotation. All images were stained following the H\&E protocol and scanned with a x20 magnification factor. 
In our experiment, we selected 225 images for training, while the remaining 25 for testing. All images are randomly cropped into 1536 × 1536 pixels patches. Similar with the CAMELYON16 dataset, we cropped and filtered 9,000 pos-patches that contained cancerous tissues 
and 18000 neg-patches, cumulatively forming the training set. From 25 testing slides, 726 pos-patches and neg-patches are included in the testing set. 

\subsection{Evaluation Protocol}
\textbf{Implementation details:}
All the experiments were implemented in PyTorch environment on a server equipped with 2 NVIDIA GeForce RTX 3090 GPUs that have 24G Memory. All image patches are resized to 256 × 256 when feeding into our model. For both dataset, Adam optimizers are employed to train the model with initial learning rate 5e-5, a weight decay 5e-4, and a batch size 16. In stage 1, the backbone would be trained for 30 epochs to get the teacher model. In stage 2, \(a=0.25\) and the teacher and student model would switch their parameters every 30 epochs. All experiments are repeated three times to calculate the mean and standard deviation. Note, in our main experiment, the simplest VGG blocks is explored as the backbone. We will substitute VGG with ResNet and other more advanced MIL-based methods in our ablation study to demonstrate the generalization of the proposal knowledge distillation framework. 

\textbf{Evaluation metrics:}
Following convention, F1-score (F1), Intersection over Union (IOU) score~\cite{everingham2015pascal}, and Hausdorff Distance(HD)~\cite{huttenlocher1993comparing} are used to evaluate the lesion segmentation performance. F1 score is the harmonic mean between precision and recall. IOU is calculated as the area of overlap between the predicted segmentation and the ground truth, divided by the area of union of the predicted segmentation and the ground truth. 
The Hausdorff Distance measures the spatial accuracy of the segmented regions, comparing the distance between the borders of the predicted segmentation and the ground truth. For each evaluation metric, we first calculate the value for each patch, and then the average of all patches is taken as the final score.

\textbf{Comparison baselines:} We applied MIL on VGG as our baseline method. For comparison, the state-of-the-art weakly supervised methods: VGG-MIL~\cite{simonyan2014very}, Resnet-MIL~\cite{he2016deep}, SA-MIL~\cite{li2023weakly}, Swin-MIL~\cite{qian2022transformer}, OAA~\cite{jiang2019integral}, PistoSeg~\cite{fang2023weakly},OEEM~\cite{li2022online} are included. VGG-MIL, Resnet-MIL, SA-MIL and Swin-MIL are MIL based methods, and OAA, OEEM and PistoSeg are CAM based pseudo supervision approaches. In addition, we conduct the evaluation on a fully supervised method CAC-Unet~\cite{zhu2021multi} for performance reference. For each method, we execute their officially-released codes on our datasets for a fair comparison.

\subsection{Results and Discussions}
\begin{table*}[t]
\scriptsize
\centering
\renewcommand{\arraystretch}{1.2}
\caption{Performance comparison to state-of-the-art weakly supervised methods, where \textbf{BOLD} highlights the best performance and$\pm$denotes the standard deviation. A full supervised approach is presented in the table for reference.}
\begin{tabular}{@{}cccccccc@{}}
\toprule
\multicolumn{1}{c|}{\multirow{2}{*}{Methods}} &\multicolumn{1}{|c|}{\multirow{2}{*}{Venues}} & \multicolumn{3}{c|}{Camelyon16}                                              & \multicolumn{3}{c}{DigestPath 2019}                                \\ \cmidrule(l){3-8} 
\multicolumn{1}{c|}{} &\multicolumn{1}{c|}{}                        & F1(\%) $\uparrow$                   & IOU(\%) $\uparrow$                & \multicolumn{1}{c|}{HD\textsuperscript{Pos} $\downarrow$}       & F1(\%) $\uparrow$                  & IOU(\%) $\uparrow$                & HD\textsuperscript{Pos} $\downarrow$                   \\ \midrule
\multicolumn{8}{l}{\textbf{Weakly Supervised Methods}}                                                                                                                           \\
\multicolumn{1}{c|}{OAA~\cite{jiang2019integral}}                   & \multicolumn{1}{c|}{CVPR'19}                   &  76.3$\pm$1.2                    &          70.5 $\pm$1.0           & \multicolumn{1}{c|}{44.5$\pm$1.1}          &     77.7$\pm$0.7                &     70.9$\pm$1.4                 &    33.5$\pm$2.8                  \\
\multicolumn{1}{c|}{OEEM~\cite{li2022online}}  & \multicolumn{1}{c|}{MICCAI'22}                 &   81.9$\pm$0.5                    &          77.8 $\pm$0.4           & \multicolumn{1}{c|}{40.9$\pm$0.9}          &     82.9$\pm$0.3                &     70.9$\pm$0.2                 &    33.6$\pm$2.1                  \\
\multicolumn{1}{c|}{PistoSeg~\cite{fang2023weakly}} & \multicolumn{1}{c|}{AAAI'23}                    &     81.2$\pm$0.3                 &     72.8 $\pm$0.9                 & \multicolumn{1}{c|}{45.8$\pm$8.7}          &   80.6$\pm$0.7                    & 72.5 $\pm$0.4                    &           29.3 $\pm$1.6        \\
\multicolumn{1}{c|}{Resnet-MIL~\cite{he2016deep}}    & \multicolumn{1}{c|}{CVPR'16}                 &     78.6$\pm$0.1                 &     72.2 $\pm$0.8                 & \multicolumn{1}{c|}{46.7$\pm$1.5}          &   74.5$\pm$0.8                    & 65.9 $\pm$0.8                    &           51.0 $\pm$3.5         \\
\multicolumn{1}{c|}{SA-MIL~\cite{li2023weakly}}  & \multicolumn{1}{c|}{MIA'23}                & 81.2$\pm$0.7          & 74.4$\pm$1.1           & \multicolumn{1}{c|}{36.9$\pm$2.8}  & 83.7$\pm$0.2          & 75.9$\pm$0.4          & 22.8$\pm$0.7            \\
\multicolumn{1}{c|}{Swin-MIL~\cite{qian2022transformer}}  & \multicolumn{1}{c|}{MICCAI'23}              & 82.1$\pm$1.1          & 77.4$\pm$1.0          & \multicolumn{1}{c|}{30.8$\pm$2.3}  &  81.8$\pm$0.2   & 75.2$\pm$0.6                  &    23.6$\pm$2.0      \\
\multicolumn{1}{c|}{VGG-MIL~\cite{simonyan2014very}}  & \multicolumn{1}{c|}{arVix'14}                   & 82.0$\pm$0.1          & 75.0$\pm$0.3          & \multicolumn{1}{c|}{27.2$\pm$0.6}  &  79.4$\pm$0.9         &      70.3$\pm$1.1        &      24.0$\pm$2.1           \\
\multicolumn{1}{c|}{Ours}    &      \multicolumn{1}{c|}{-}          & \textbf{85.6$\pm$0.1}       & \textbf{80.0$\pm$0.1}           & \multicolumn{1}{c|}{\textbf{19.0$\pm$0.4}} &   \textbf{85.4$\pm$0.1}               &  \textbf{77.7$\pm$0.1}             &  \textbf{14.0$\pm$0.8}               \\ \midrule
\multicolumn{8}{l}{\textbf{Fully Supervised Method}}                                                       \\
\multicolumn{1}{c|}{CAC-Unet~\cite{zhu2021multi}}    & \multicolumn{1}{|c|}{NeuCompt'21}            &     88.3$\pm$0.1                 &     84.5$\pm$0.1                 & \multicolumn{1}{c|}{16.7$\pm$0.4}          &  90.1$\pm$0.1                  &           85.5$\pm$0.1           &        9.6$\pm$0.4              \\ \bottomrule
\end{tabular}
\label{table:2}
\end{table*}

\begin{figure*}[!ht]
\centering
    \includegraphics[width=0.95\textwidth]{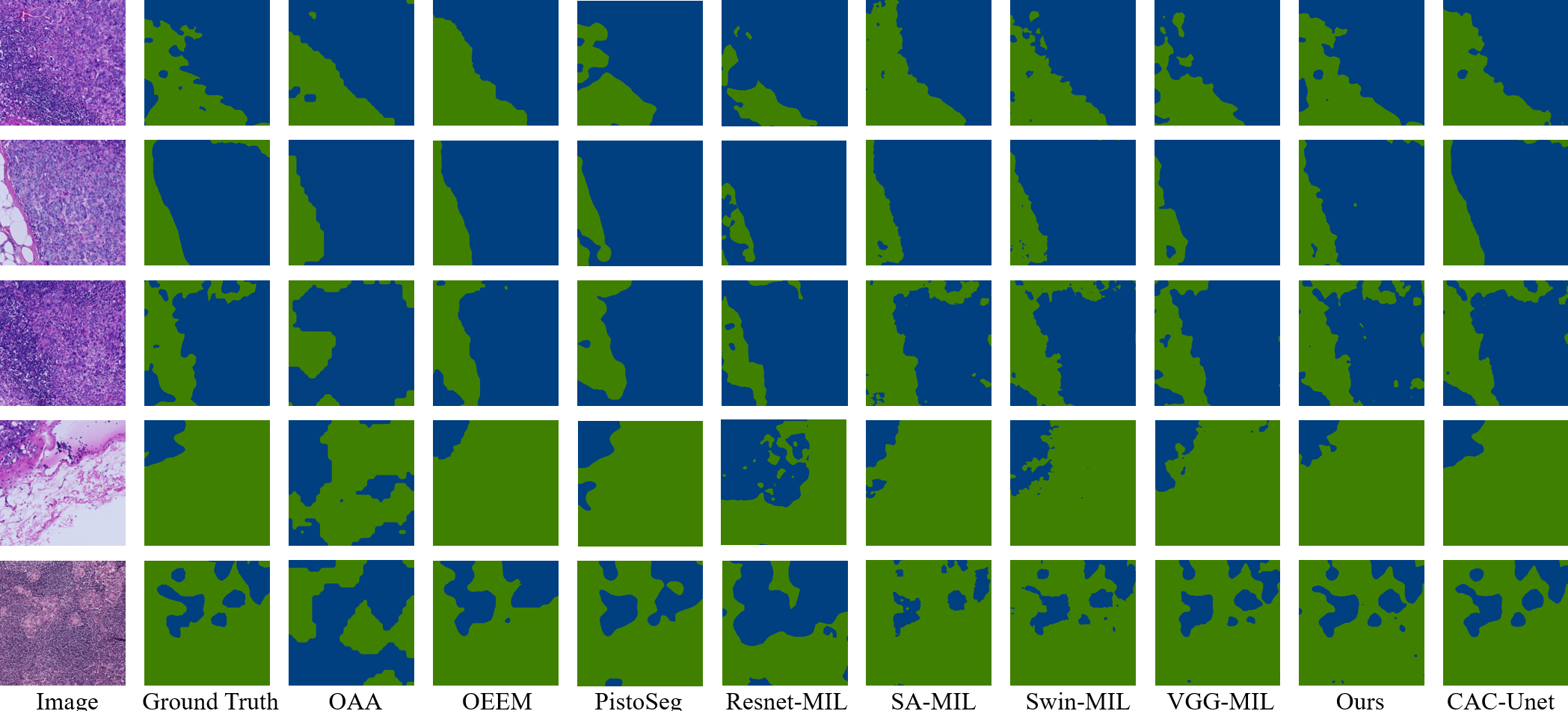}
    \caption{Visualization of segmentation results on the Camelyon16. The cancer tissues are shown in blue, and normal tissues are shown in green. OAA~\cite{jiang2019integral}, OEEM~\cite{li2022online}, and PistoSeg~\cite{fang2023weakly} follow the CAM-based pseudo-supervision paradigm, and SA-MIL~\cite{li2023weakly}, Swin-MIL~\cite{qian2022transformer}, Resnet-MIL~\cite{he2016deep}, and our baseline VGG-MIL are MIL-based approaches. For reference, we also include the results by a fully-supervised approach CAC-Unet~\cite{zhu2021multi}.}
    \label{fig:results_vis}
\end{figure*}

\textbf{Main results:} The quantitative evaluation results are presented in Table \ref{table:2}. Our approach surpasses SOTA weakly-supervised learning methods across both datasets. Particularly, compared to the VGG-MIL baseline, the proposed iterative knowledge distillation strategy brings in significant performance boosts. The qualitative evaluation are displayed in Fig. \ref{fig:results_vis}. Weakly-supervised learning methods based on Multiple Instance Learning tend to generate substantial noise in their predictions. In contrast, methods reliant on Class Activation Mapping (CAM) and pseudo-labeling struggle to produce sufficiently accurate segmentation maps. By integrating the strengths of both approaches, our model has achieved significant improvements in terms of accuracy and stability. 

\subsubsection{Compatible to MIL-based methods:}
As we previously stated, the backbone of our model is not fixed and our iterative knowledge distillation strategy can be easily adapted to different MIL-based methods. To substantiate our claim, we substituted the model backbone with Resnet, SA-MIL and Swin-MIL and conducted tests on both datasets and report the experiment results in Table \ref{table:3}. Our iterative fusion-knowledge distillation method boosts the performance of all 4 MIL-based segmentation methods with considerable large margins.

\begin{table}[t]
\small
\centering
\renewcommand{\arraystretch}{1.2}
\caption{Applying our iterative fusion-knowledge distillation (IFKD) to different MIL-based segmentation methods. Substantial performance improvements is observed. }
\begin{tabular}{c|cccccc}
\toprule
&\multicolumn{3}{|c|}{Camelyon16}&\multicolumn{3}{c}{Digestpath2019}\\
Methods  &F1(\%) $\uparrow$                   & IOU(\%) $\uparrow$                & \multicolumn{1}{c|}{HD\textsuperscript{Pos} $\downarrow$}       & F1(\%) $\uparrow$                  & IOU(\%) $\uparrow$                & HD\textsuperscript{Pos} $\downarrow$ 

\\ \midrule
\multirow{1}{*}{VGG-MIL}
& 82.0$\pm$0.1          & 75.0$\pm$0.3          & \multicolumn{1}{c|}{27.2$\pm$0.6}  &  79.4$\pm$0.9         &      70.3$\pm$1.1        &      24.0$\pm$2.1 \\
+ FKSD  
& {85.6$\pm$0.1}       & {80.0$\pm$0.1}           & \multicolumn{1}{c|}{{19.0$\pm$0.4}} &   {85.4$\pm$0.1}               &  {77.7$\pm$0.1}             & {14.0$\pm$0.8} 

\\ \midrule
\multirow{1}{*}{ResNet-MIL}
& 78.6$\pm$0.1 & 72.2$\pm$0.8 & \multicolumn{1}{c|}{46.7$\pm$1.5} & 74.5$\pm$0.8 & 65.9$\pm$0.8 & 51.0$\pm$3.5\\
+ FKSD  
& 83.4$\pm$0.1 & 77.8$\pm$0.1 & \multicolumn{1}{c|}{33.6$\pm$0.3}
& 83.2$\pm$0.2 & 75.5$\pm$0.2& 22.6$\pm$0.5
 
 \\ \midrule
\multirow{1}{*}{Swin-MIL}     
& 82.1$\pm$1.1 & 77.4$\pm$1.0 & \multicolumn{1}{c|}{30.8$\pm$2.3}
& 81.8$\pm$0.2 & 75.2$\pm$0.6& 23.6$\pm$2.0
\\
+ FKSD 
& 84.4$\pm$0.2 & 78.0$\pm$0.2 & \multicolumn{1}{c|}{23.5$\pm$0.4}
& 83.5$\pm$0.3 & 76.5$\pm$0.2& 23.5$\pm$0.6
 
\\ \midrule
\multirow{1}{*}{SA-MIL}     
& 81.2$\pm$0.7 & 74.4$\pm$1.1 & \multicolumn{1}{c|}{36.9$\pm$2.8}
& 83.7$\pm$0.2 & 75.9$\pm$0.4& 22.8$\pm$0.7
\\
+ FKSD 
& 85.1$\pm$0.1 & 79.1$\pm$0.1 & \multicolumn{1}{c|}{21.2$\pm$0.3}
& 84.1$\pm$0.2 & 76.8$\pm$0.1& 20.1$\pm$0.8\\

\bottomrule
\end{tabular}
\label{table:3}
\end{table}

\begin{figure}[!ht]
    \centering
    \includegraphics[width=\textwidth]{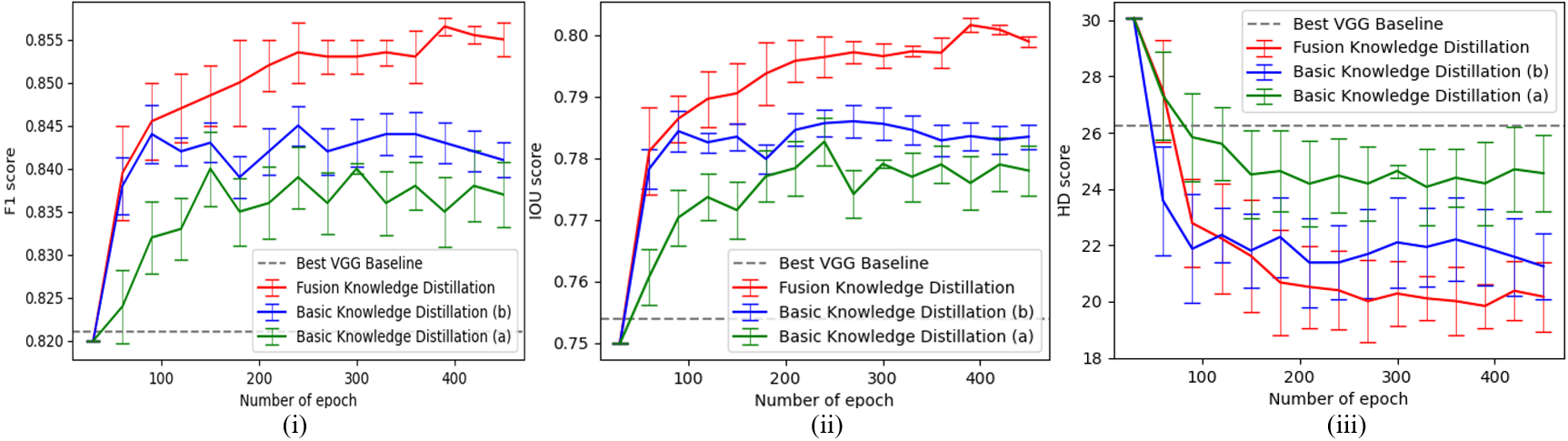}
    \caption{Ablation study for fusion-knowledge distillation on Camelyon16. Each point represents the best result in the past 30 epochs, and the standard deviation is marked by the error bars. The dashed line is the best score that teacher model could achieve during MIL. Basic knowledge distillation (a) and (b) correspond to the knowledge distillation structures shown in Fig.~\ref{fig:skd}.}
    \label{fig:Camelyon16_fusion}
\end{figure}
\begin{figure}[!ht]
    \centering
    \includegraphics[width=\textwidth]{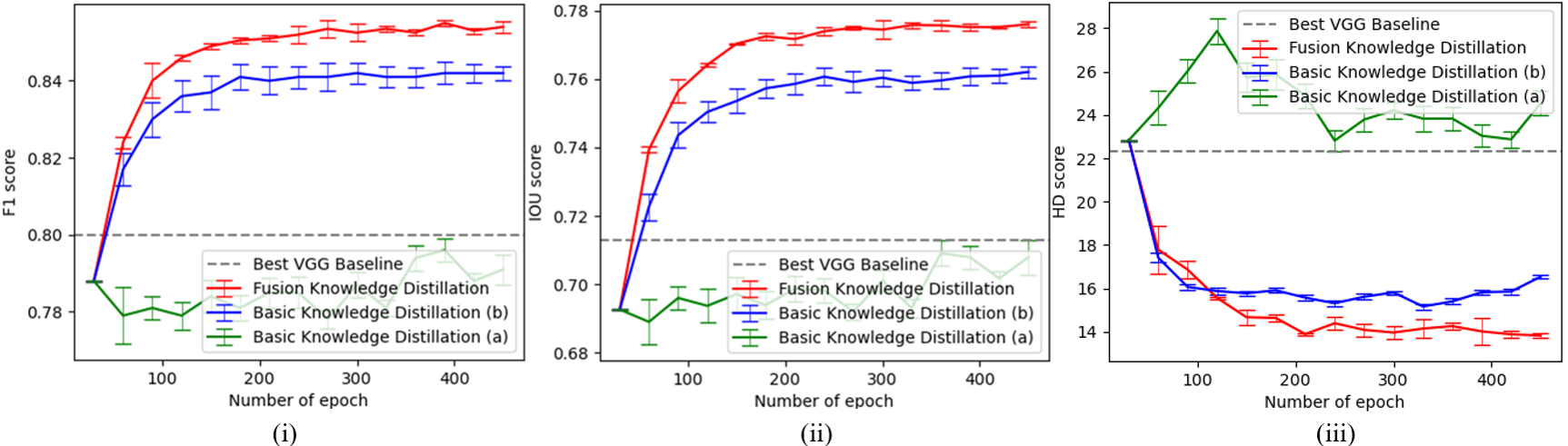}
    \caption{Ablation study for fusion-knowledge distillation on Digestpath2019.}
    \label{fig:Digestpath2019_fusion}
\end{figure}

\subsection{Ablation Studies}
Ablations are performed to assess the effectiveness of the building blocks in our solution. Each test was repeated for 3 rounds, and mean and standard deviation were reported. All ablation studies in this section takes VGG as the backbone.

\begin{figure}[t]
    \centering
    \includegraphics[width=\textwidth]{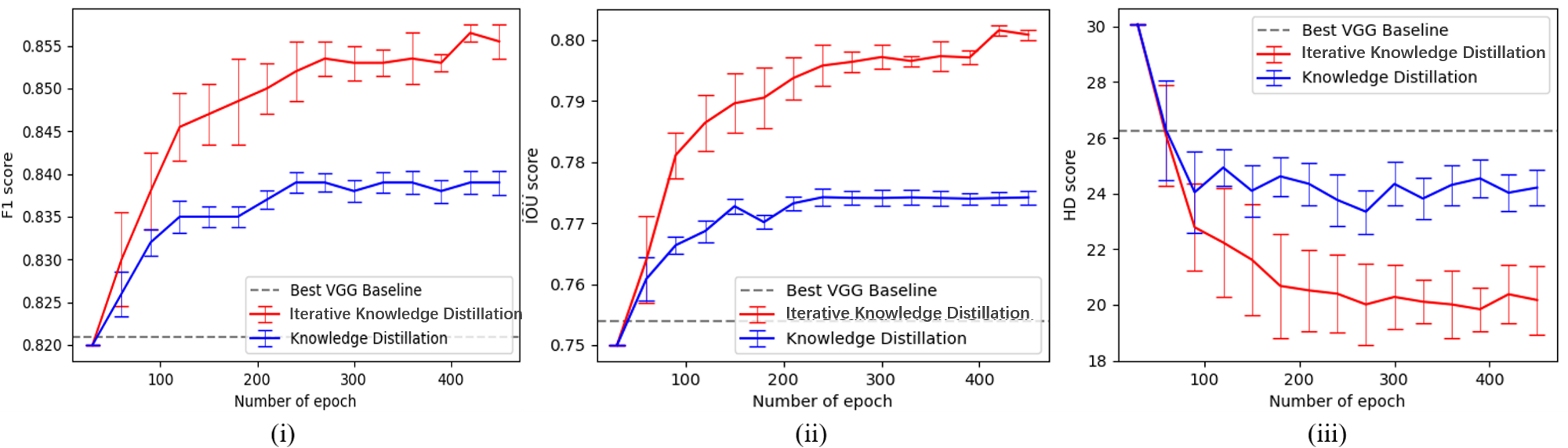}
    \caption{Ablation on teacher-student role switch in iterative fusion-knowledge distillation on Camelyon16. Each point represents the best result in the past 30 epochs, and the standard deviation is marked by the error bars. The dashed line is the best performance that teacher model could achieve during MIL.}
    \label{fig:Camelyon16_SKD}
\end{figure}
\begin{figure}[t]
    \centering
    \includegraphics[width=\textwidth]{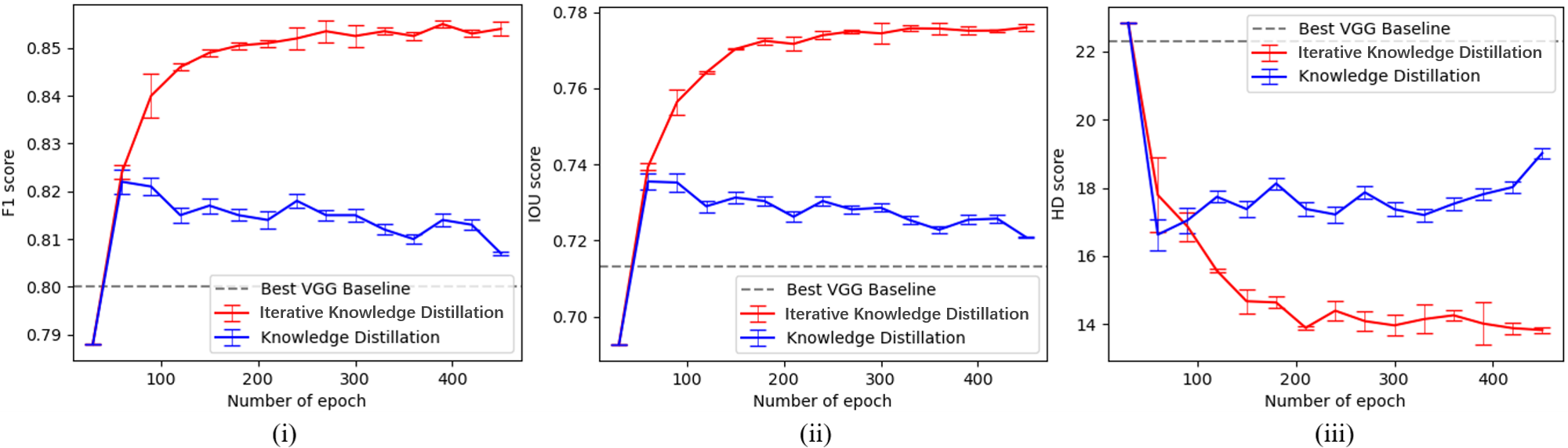}
    \caption{Ablation on teacher-student role switch in iterative fusion-knowledge distillation on Digestpath2019. }
    \label{fig:Digestpath2019_SKD}
\end{figure}

\begin{table}[t]
\small
\centering
\renewcommand{\arraystretch}{1.2}
\caption{Ablation on our iterative fusion-knowledge distillation objective function.}
\begin{tabular}{ll|lll|lll}
\midrule
 & & \multicolumn{3}{c|}{Camelyon16}&\multicolumn{3}{c}{Digestpath2019}  \\\midrule
\multicolumn{1}{c}{$L_{\text{kd}}$} & \multicolumn{1}{c|}{$L_{\text{wce}}$} & \multicolumn{1}{c|}{F1}         & \multicolumn{1}{c|}{IOU}        & \multicolumn{1}{c|}{HD} & \multicolumn{1}{c|}{F1}         & \multicolumn{1}{c|}{IOU}        & \multicolumn{1}{c}{HD} \\\midrule
 \checkmark &  &\multicolumn{1}{l|}{85.1$\pm$0.2}  & \multicolumn{1}{l|}{ 78.9$\pm$0.2}    & 22.1$\pm$1.0       &  \multicolumn{1}{l|}{84.5$\pm$0.5}    &   \multicolumn{1}{l|}{76.4$\pm$0.4}    &      15.5$\pm$1.4       \\
 
 \checkmark&\checkmark&\multicolumn{1}{l|}{85.6$\pm$0.1}    &  \multicolumn{1}{l|}{80.0$\pm$0.1}    & 19.0$\pm$0.4        &  \multicolumn{1}{l|}{85.4$\pm$0.1}    &   \multicolumn{1}{l|}{77.1$\pm$0.1}    &  14.0$\pm$0.8  \\
 \midrule
\end{tabular}
\label{table: loss ablation}
\end{table}

\begin{table}[t]
\small
\centering
\renewcommand{\arraystretch}{1.2}
\caption{Ablation on the hyperparameter $a$ of the objective function on Camelyon16.}
\begin{tabular}{c|c|c|c|c|c|c}
\hline
\  $a$            & 0    & 0.1  & 0.25 & 0.5  & 1    & 5    \\ \hline
\  F1 Score(\%) \ &\  85.1 \ & \ 85.5 \  &\  85.6 \ &\  85.2 \  & \ 84.2 \  & \ 81.0 \  \  \\ \hline
\end{tabular}
\label{table: a ablation}
\vspace{-10pt}
\end{table}

\noindent\textbf{Ablation on Fusion-knowledge Distillation:}
We compare our fusion-knowledge distillation to conventional knowledge distillation paradigms shown in Fig. 1 and report their performance over 450 epochs on Camelyon16 and Digetpath2019 in Fig.~\ref{fig:Camelyon16_fusion} and \ref{fig:Digestpath2019_fusion}, respectively. For basic knowledge distillation paradigms (a) and (b), the shallow layers of the student model are prone to learning local information from the teacher, while student's deeper layers can only learn global semantic information. Consequently, the improvement to the student model's performance is relatively limited. In contrast, with fusion knowledge distillation, each block of the student model learns from the multi-dimensional features extracted by the teacher model, which empowers student's segmentation capability. It is noteworthy that on the Digestpath2019 dataset, the performance of the knowledge distillation method (a) falls below the baseline of the VGG model. We speculate that the performance drop arises from a lack of constraints on the early layers of the student model within the knowledge distillation process.
This also highlights the necessity of fusing segmentation information from different dimensions.



\noindent\textbf{Ablation on Teacher-Student Role Switch in Iterative Fusion-knowledge Distillation:}
Our iterative fusion-knowledge distillation strategy proposes to iteratively switch the role between the teacher and student models. That is, the teacher model is also updated. In this ablation, we compare the segmentation performance with and without the teacher update and show the results in Fig.~\ref{fig:Camelyon16_SKD} and Fig.~\ref{fig:Digestpath2019_SKD}. With the same teacher-student framework, no teacher update tends to stabilize in performance around 200 epochs, but our iterative fusion-knowledge distillation continues to bolster overall performance by updating the teacher model. 

\noindent\textbf{Ablation on Iterative Fusion-knowledge Distillation Objective Function and Hyper-parameter:}
Our iterative fusion-knowledge distillation strategy is harnessed by two loss terms: $L_{kd}$,  $L_{wce}$. Table \ref{table: loss ablation} shows their impacts on segmentation results on both Camelyon16 and Digestpath2019. $L_{wce}$ substantially improves the network's segmentation capability and stability. The two loss terms are combined by a hyperparameter $a$ in our method. Table \ref{table: a ablation} quantifies sensitivity of the solution to its numerical setting. It shows that the system is quite stable when $a<1$.


\section{Conclusion \& Limitation}
In this study, we introduced a novel MIL-based iterative fusion-knowledge distillation framework for weakly supervised semantic segmentation in histopathology. The iterative fusion-knowledge distillation strategy significantly enhances MIL-based segmentation by addressing noise and label ambiguity, enabling a deeper and more comprehensive learning from the teacher model's outcomes. 
Extensive experiments on Camelyon16 and DigestPath 2019 datasets show that our approach surpasses SOTA methods by a large margin. 

It should be noted that the process of iterative fusion-knowledge distillation, especially with the dynamic role reversal and continuous refinement of models, can be computationally intensive. This may require significant computational resources and time, potentially limiting its applicability in resource-constrained scenarios.

\bibliographystyle{splncs04}
\bibliography{main}
\end{document}